\documentclass{article} 
\usepackage{nips14submit_e,times}
\usepackage{url}
\usepackage{graphicx}
\usepackage{array}
\usepackage{amssymb,amsmath,bm} 
\usepackage{stmaryrd} 
\usepackage[boxed,ruled,vlined,linesnumbered]{algorithm2e}
\usepackage{multirow} 
\usepackage{makecell} 
\usepackage{arydshln} 
\usepackage{subfigure}
\usepackage[colorlinks, linkcolor=red,anchorcolor=green, citecolor=blue]{hyperref}

\title{Multi-layered Semantic Representation Network for Multi-label Image Classification}

\author{
Xiwen Qu$^1$, Hao Che$^{3,*}$, Jun Huang$^{1,2}$\thanks{chehao17@gmail.com, huangjun.cs@ahut.edu.cn}, Linchuan Xu$^{3}$, Xiao Zheng$^{1,2}$ \\\\
$^1$School of Computer Science and Technology, Anhui University of Technology, China\\
$^2$Institute of Artificial Intelligence, Hefei Comprehensive National Science Center, China\\
$^3$Australian National University, Australia\\
$^4$Department of Computing,The Hong Kong Polytechnic University, Hong Kong SAR, China\\
}

\nipsfinalcopy 

\begin{document}

\maketitle

\begin{abstract}
Multi-label image classification (MLIC) is a fundamental and practical task, which aims to assign multiple possible labels to an image.
In recent years, many deep convolutional neural network (CNN) based approaches have been proposed which model label correlations to discover semantics of labels and learn semantic representations of images. This paper advances this research direction by improving both the modeling of label correlations and the learning of semantic representations. On the one hand, besides the local semantics of each label, we propose to further explore global semantics shared by multiple labels.
On the other hand, existing approaches mainly learn the semantic representations at the last convolutional layer of a CNN.
But it has been noted that the image representations of different layers of CNN capture different levels or scales of features and have different discriminative abilities. We thus propose to learn semantic representations at multiple convolutional layers. To this end, this paper designs a Multi-layered Semantic Representation Network (MSRN) which discovers both local and global semantics of labels through modeling label correlations and utilizes the label semantics to guide the semantic representations learning at multiple layers through an attention mechanism.
Extensive experiments on four benchmark datasets including VOC 2007, COCO, NUS-WIDE, and Apparel show a competitive performance of the proposed MSRN against state-of-the-art models.

\end{abstract}

\section{Introduction}
\label{sec:intro}
Multi-label image classification (MLIC) deals with assigning multiple labels to each image,
and it has been applied in many fields, including multi-object
recognition \cite{CVPR2016:ODVT}, medical diagnosis recognition \cite{Chest-X-rays} and Person re-identification \cite{CVPR2020:PRID}. The recent progress is mainly made by exploiting label correlations and learning semantic representations with deep learning models.

Modeling label correlations has been long studied in multi-label classification and has been demonstrated very effective because correlated labels are highly likely to co-occur \cite{CVPR2019:ML-GCN}. For an image recognition task, deep learning models, especially convolutional neural networks (CNNs), have been widely recognized as state-of-the-art models. Hence, many recent approaches to MLIC are based on deep learning models and exploiting label correlations, e.g., \cite{CVPR2019:ML-GCN,AAAI2018:OFRNN,ICME2018:RCDH}.
In these approaches, an existing deep learning model is usually employed as a tool to transform a raw image into a high-level abstract representation. But objects of interest may only be in certain regions of an image. Therefore, some studies \cite{ICCV2017:RDAR,CVPR2017:LSRIS,AAAI2018:RARL,ICCV2019:ssgrl,AAAI2020:CMA} utilize semantics of labels to guide the learning of semantic representations of images. The semantic representations are expected to only cover the regions of interest, and thereby improve the classification performance.

This paper advances this research direction by improving both the modeling of label correlations and the learning of semantic representations. On the one hand, unlike existing approaches only learn local semantics of each label, we propose to further explore global semantics shared by multiple labels. The global semantics can effectively capture high-order correlations among multiple labels. On the other hand, existing approaches mainly learn the semantic representations at the last convolutional layer of a CNN.
But it has been noted that the image representations at different layers of CNN capture different levels or scales of features and have different discriminative abilities \cite{ECCV2016:UMDC,ICLR2018:MLAtt,DenseNet,ICCV2015:MSR}. Therefore, better performance might be achieved by simultaneously exploiting features at multiple layers of a CNN.

To realize the proposals mentioned above, we design a novel Multi-layered Semantic Representation Network (MSRN). MSRN generates both label-specific and group embeddings which capture local semantics and global semantics respectively,
and then combines them with multiple layers of a CNN by an attention mechanism to learn label and group shared semantic representations of images.
To be specific, first, we introduce LGE (Label-Group Embedding) module to capture both local semantics of each label and semantics of a group of labels in embeddings based on the label co-occurrence graph.
Second, we propose SGA (Semantic Guided Attention) module to explicitly guide the CNN to focus on the regions of interest.
Third, we design a framework to combine the LGE module with the multiple layers of
a CNN through the attention mechanism built in the SGA module.
We conduct experiments on four benchmark multi-label image datasets including COCO, VOC 2007, Apparel and NUS-WIDE.
The experimental results show that our method outperforms state-of-the-art approaches.

The contributions of this paper are summarized as follows:
\begin{itemize}
    \item A Multi-layered Semantic Representation Network (MSRN) is designed for multi-label image classification.
    \item Second-order and high-order label correlations are considered simultaneously to improve the performance of multi-label image classification.
    \item Semantic representations are learned at multiple layers through an attention mechanism by modeling label correlations.
\end{itemize}

The rest of the paper is organized as follows. Section \ref{sec:relatedwork} introduces related work. Section \ref{sec:method} presents the proposed method. Section \ref{sec:experiment} presents empirical evaluation.
Section \ref{sec:conclusion} concludes this paper and introduces future work.

\section{Related Work}
\label{sec:relatedwork}
In MLIC, images are annotated with multiple labels simultaneously where labels usually have correlations.
It has been demonstrated that exploiting label correlations can significantly improve the performance \cite{TKDE14:MLZhang:MLReview}.
Recent progress has been made by employing deep learning models, especially convolutional neural networks. Wang et al. \cite{CVPR2016:ULMIC} extract label semantics and associate it to Recurrent Neural Network (RNN).
In addition,
Lee et al. \cite{CVPR2018:ZSL} apply knowledge graphs to exploit the label dependencies based on the
label co-occurrence graph.
ML-GCN \cite{CVPR2019:ML-GCN} learns the semantic label embeddings through Graph
Convolution Network (GCN),
and applies it as inter-dependent object classifiers at the prediction stage.
In \cite{AAAI2020:KSSNet}, a label graph superimposing framework is proposed to exploit label correlations. The label graph is constructed by superimposing statistical label graph into knowledge prior oriented graph, which, however, is usually unavailable in real applications.

Some studies further locate regions of interest because each class label might be determined by some specific regions of an image. Examples include \cite{ECCV2014:EB} \cite{CVPR2016:EBBA} which apply bounding box to focus on the regions of proposal. To learn regions with arbitrary boundaries, more studies propose attention based methods where attention is a spatial weight map representing relative importance among pixels \cite{ICLR2018:MLAtt}.
SRN \cite{CVPR2017:LSRIS} is an end-to-end CNN model
which trains learnable convolutions on the attention maps of labels.
In \cite{AAAI2018:OFRNN}, Chen et al. propose an order-free RNN based model
for multi-label image classification, which uniquely integrates the learning of visual attention and
Long Short Term Memory (LSTM) layers to jointly learn the labels of interest and their co-occurrences.

Recently, some methods \cite{AAAI2020:CMA} \cite{CVPR2019:ML-GCN} \cite{ICCV2019:ssgrl} apply Graph
Neural Network (GNN) techniques to generate semantic label embeddings which can be utilized as visual
attention for multi-label image classification.
You et al. \cite{AAAI2020:CMA} propose a method of computing cosine similarity
between label embeddings to exploit label dependencies.
Chen et al. \cite{ICCV2019:ssgrl} apply a GNN with graph propagation mechanism
to exploit the interaction between DNN and label dependencies.
Despite having achieved high performance on multi-label image classification,
these methods do not explicitly consider high-order label dependencies which may result in semantics shared by a group of labels \cite{TKDE14:MLZhang:MLReview}.
Moreover, to the best of our knowledge, no existing approaches utilize image representations extracted from multiple layers of a CNN.

\section{Method}
\label{sec:method}
\subsection{Architecture}
\begin{figure}[tb]
\centering
\includegraphics[width=0.7\textwidth]{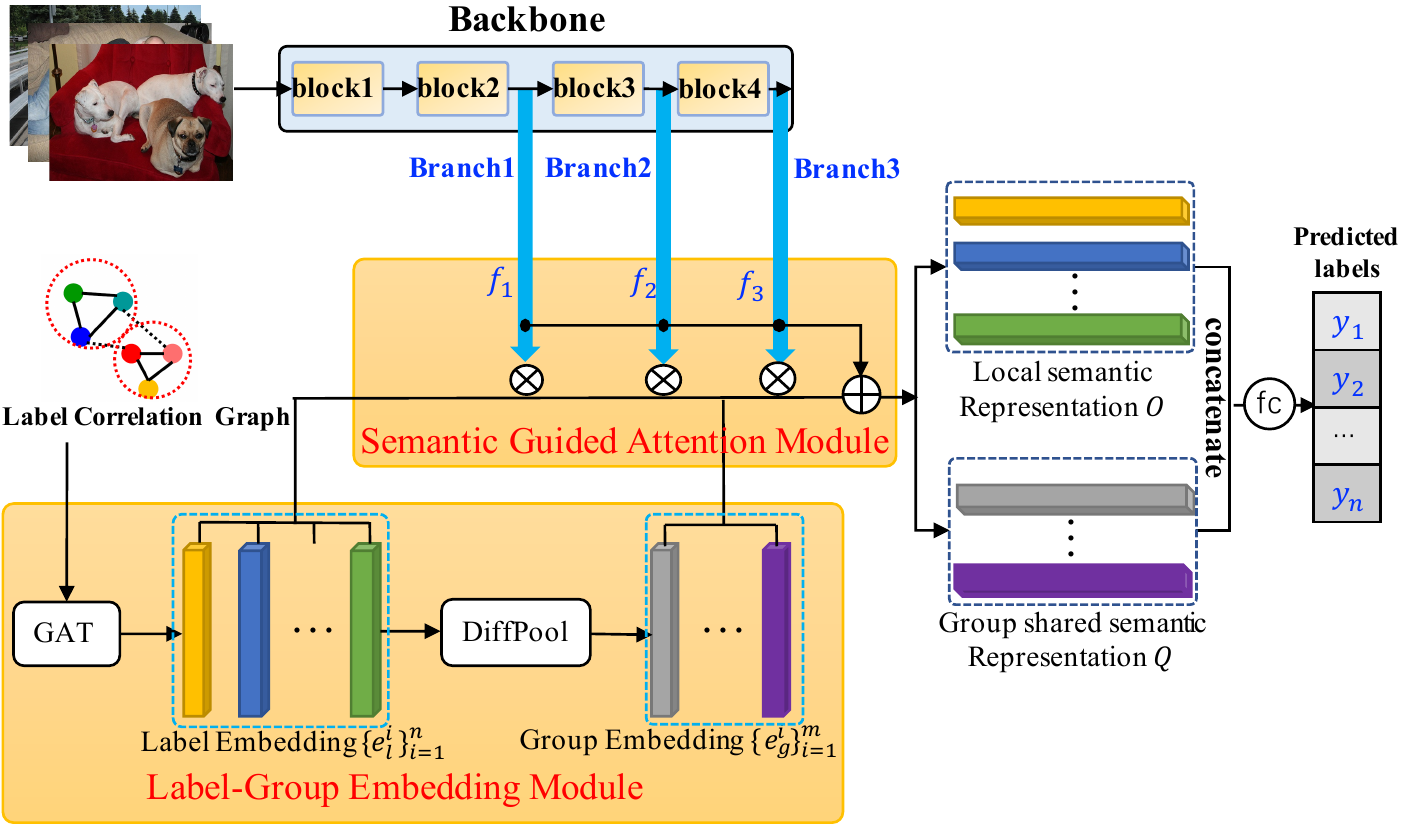}
\caption{Overall architecture of our MSRN model. Given an image, a CNN network outputs the image features from different layers to different branches.
At the same time, we input label correlation graph to LGE module which can output the label and the group embeddings capturing label semantics and group semantics, respectively.
Then, the SGA module produces label-level semantic representations and group-level semantic representations of the image by combining each
image feature map from each branch with the learned label and group embeddings through an attention mechanism.
At last, we concatenate the generated semantic representations and apply fully connected layers to perform the classification.
}
\label{fig:2}
\end{figure}

The architecture of the proposed MSRN is shown in Figure \ref{fig:2}.
We design an LGE module to generate label and group embeddings with the input of a graph $\mathcal{G} = \{V, A\}$. $V = \{v_{i}\}_{i=1}^n$ is the feature matrix of labels where $v_{i}$ is a vector of features and $n$ is the number of labels,
and ${A} = \{a_{ij}\}_{i,j=1}^n$ is the adjacency matrix about label co-occurrence.
The outputs of LGE module are label embeddings $E_l= \{e_l^i\}_{i=1}^n \in \mathbb{R}^{n\times d}$
and group embeddings $E_g = \{e_g^i\}_{i=1}^m\in \mathbb{R}^{m\times d}$, where $m$ is the number of groups, and $d$ is the dimension of the embeddings.

The backbone in our framework can be any kind of CNNs,
such as VGG \cite{Vgg}, ResNet \cite{ResNet} and DenseNet \cite{DenseNet}.
In this paper, Resnet-101 is chosen for experiment.
Given an input image, $\overline{F} = \{\overline{f}_{\emph b}\}_{\emph i=1}^\emph B$ indicates the output image features of different branches,
where $B$ is the total number of branches, and ${\overline{f}_{\emph b}} \in \mathbb{R}^{\emph W_\emph b\times \emph H_\emph b\times \emph C_\emph b}$
is image feature for the $b$-th branch with spatial resolution $W_\emph b\times H_\emph b$ and channel $C_\emph b$.
The branches marked as blue lines in Figure \ref{fig:2}
are used to receive the image feature map $\overline{f}_{\emph b}$ from corresponding layers in the CNN.
We provide three branches in our work to receive the output image features from the last layer of the last three blocks of Resnet-101.
Since the channel $C_\emph b$ of image features from different layers of CNN are distinct,
we use a convolutional layer with 1$\times$1  kernel to project image features from
${\overline{f}}_\emph b$ to ${f_{\emph b}} ={conv}^{ 1\times 1}(\overline{f}_\emph b)\in \mathbb{R}^{\emph W_\emph b\times \emph H_\emph b\times \emph d}$
which has the same dimension $d$ as the label embeddings $E_l$ and group embeddings $E_g$.

Then we propose an SGA module to position-wisely combine
the image feature maps $F = \{{f}_{\emph b}\}_{\emph i=1}^\emph B$ with
label embeddings $E_l$ and group embeddings $E_g$.
The outputs of SGA module are label semantic representations $O =\{{o}_{\emph b}\}_{\emph b=1}^\emph B$
and group shared semantic representations $Q= \{{q}_{b}\}_{b=1}^B$,
where $o_b\in \mathbb{R}^{n\times d}$ and $q_b\in \mathbb{R}^{m\times d}$.
Some of the existing approaches utilize the label-specific representation for each label. But since in real applications the labeling results of a dataset usually have noisy or missing labels, the label-specific semantic representation might not be sufficient enough to predict correct labels. Therefore, in the final stage of our framework,
we concatenate the generated label and group shared semantic representations into $M=[O||Q]$, and apply fully connected layers to perform the prediction where the cross entropy loss function is adopted as follows:
\begin{eqnarray}
\mathcal{L}_{1}= \sum_{i=1}^{n}y^{i}\log(\sigma(\hat{y}^{i})) + (1-y^{i})\log(1-\sigma(\hat{y}^{i})),
\label{sec:bceloss}
\end{eqnarray}
where $y^{i}$ is equal to 1 or 0 for image $i$ in terms of a certain label, $\hat{y}^{i}$ is the output of fully connected layer, and $\sigma(\cdot)$ is the sigmoid function.

\subsection{Label-Group Embedding Module}
Since label correlation is important information in multi-label
image classification as we mentioned in section \ref{sec:intro},
we build a Label-Group Embedding (LGE) module to generate semantic label embeddings $E_l$ and group embeddings $E_g$.

\subsubsection{Semantic Label Embeddings}
Graph attention Networks (GAT) \cite{ICLR2018:GAT} is a self-attention based model which is most frequently used for learning embeddings of graph-structured data.
With GAT algorithm, we can obtain the semantic label embeddings $E_l$ from the label graph $\mathcal{G}$.
In our model, GAT first produces the attention coefficient $\alpha_{{ij}}$ between the $i$-th and $j$-th label as follows:
\begin{eqnarray}
\alpha_{{ij}} = \frac{\exp(\text{LeakyReLU}(P[{U}v_{i}||{U}v_{j}]))}{\sum_{k \in \mathcal{N}_i} \exp(\text{LeakyReLU}(P[{U}v_{i}||{U}v_{k}]))},
\label{sec:gat1}
\end{eqnarray}
where $P \in \mathbb{R}^{1 \times 2w}$ and $U\in \mathbb{R}^{w\times v}$ are two learnable weight matrices,
$\mathcal N_i$ is the set of neighborhoods of label $i$ in the graph,
and $||$ represents the concatenation operation.
The negative input slope in LeakyReLU is set to be 0.2 in our work.
Then, we can obtain label embeddings $E_l^1 = \{e_l^i\}_{i=1}^n$ from the first GAT layer
by linearly combining attention coefficients $\alpha$ with the transformed label features:
\begin{eqnarray}
e_l^i =\sigma(\sum_{j \in \mathcal{N}_i}\alpha_{{ij}}{U}v_{j}  + {U}v_{i}),
\label{sec:gat2}
\end{eqnarray}
where the $\sigma(\cdot)$ is non-linear activation function which is ELU in our method.
For simplicity, $\text{GAT}_t(\cdot)$ is used to represent the $t$-th
GAT layer that consists of equations (\ref{sec:gat1}) and (\ref{sec:gat2}),
and the semantic label embeddings $E_l^t$ can be generated by the following equation:
\begin{eqnarray}
E_l^t = \text{GAT}_{t}(E_l^{t-1}, A) \label{con:label},
\end{eqnarray}
where $E_l^0=V$ is the original feature matrix of labels.

\subsubsection{Semantic Group Embeddings}
Differentiable graph pooling (Diffpool)\cite{NIPS18:Diffpool} is a graph clustering algorithm
that soft map graph nodes to a set of clusters.
Once we capture the semantic label embeddings $E_l$, we can apply Diffpool to generate semantic group embeddings $E_g$ as
\begin{eqnarray}\label{diffpool2}
E_g =\text{Diffpool}(E_l,A).
\end{eqnarray}

Moreover, in order to learn more compact group embeddings,
we try to minimize the distance between the group embeddings $E_g$ and the labels embeddings $E_l$ as follow
\begin{eqnarray}\label{sec:grouploss}
\mathcal{L}_{2} = \sum_{k=1}^{m} \sum_{ E_l^{i}\in C_{k}} \| E_g^{k} - E_l^{i}\|_2^{2},
\end{eqnarray}
where $C_{k}$ indicates the $k$-th cluster of labels which are highly correlated labels.

\subsection{Semantic Guided Attention Module}

The aim of SGA module is
to utilize the semantic embeddings $E_l$ and $E_g$ to guide the
learning of semantic representations of images at different
branches. As the feature contained in each position (w, h)
of an image feature map could be correlated to the semantics
of the label embeddings, we propose a position-wise attention
mechanism to fully combine the image feature space and the
semantic embedding space. Similar to existing studies \cite{ICLR2017:HPLBP, ICCV2019:ssgrl},
we adopt the Hadamard product between each position $(w, h)$ of an image feature map from the $b$-th branch and the label, group
embeddings to calculate the attention weights as
\begin{eqnarray}
{sl_{b}}_{w,h}^i=f^{w,h}_{b} \odot e_l^i,~~~
{sg_{b}}_{w,h}^j=f^{w,h}_{b} \odot e_g^j,
\end{eqnarray}
where the $\odot$ is Hadamard product, ${sl_{b}}_{w,h} \in \mathbb{R}^{1 \times 1 \times n \times d}$ and ${sg_{b}}_{w,h} \in \mathbb{R}^{1 \times 1 \times m \times d}$. Then we apply normalization to the computed compatibility scores $al_{b} \in \mathbb{R}^{W_b \times H_b \times n \times d}$ and $ag_{b} \in \mathbb{R}^{W_b \times H_b \times m \times d}$
\begin{eqnarray}\label{con: normalized_group}
al^{w,h}_{b} = \frac{\exp({sl_{b}}_{w,h} )}{\sum_{x,y}\exp({sl_{b}}_{x,y})},
ag^{w,h}_{b} = \frac{\exp({sg_{b}}_{w,h} )}{\sum_{x,y}\exp({sg_{b}}_{x,y})}.
\end{eqnarray}

Once obtained the normalized compatibility scores,
we apply the second Hadamard product to generate position-wise attention maps.
\begin{eqnarray}\label{con:normscore}
o_{b} = \sum_{w,h}al^{w,h}_{b} \odot f^{w,h}_{b}, ~~~q_{b} = \sum_{w,h}ag^{w,h}_{b} \odot f^{w,h}_{b}.
\end{eqnarray}

Finally, we concatenate the local semantic representations $O$$=$$\{o_b\}_{b=1}^B$$\in$$\mathbb{R}^{n\times (Bd)}$
and group shared semantic representations
$Q$$=$$\{q_b\}_{b=1}^B$$\in$$\mathbb{R}^{m\times (Bd)}$
and predict the labels
by $\hat{y}^{i}$=${fc}_{2}(\text{LeakeyReLU}({fc}_{1}(\text{tanh}(M))))$, where $M$=$[O||Q]$, ${fc}_{1}$ and ${fc}_{2}$ are fully connected layers.
The total training loss is $\mathcal{L}_{1} + \lambda \mathcal{L}_{2}$,
where $\lambda$ is a regularization parameter.

\section{Empirical Evaluation}
\label{sec:experiment}
In this section, we will describe the implementation details of our proposed model MSRN and the experimental results.

\subsection{Implementation Details and Evaluation Metrics}
The input label features $V$ are 300-dimensional Glove features pretrained on Wikipedia dataset.
The backbone ResNet-101 is pretrained on ImageNet for accelerating
training process. We remove the last average pooling layer and classifier of Resnet-101
and apply the MaxPooling with kernel size $2\times 2$ and stride 2 to obtain image features, $\overline{F}$,
from last three building blocks of ResNet-101. The output dimension of ${fc}_{1}$ is 2048, and the output dimension of ${fc}_{2}$ is the same as the number of labels. In addition, to reduce the impact of branches
corresponding to lower layers of the backbone on gradients, we add a buffer convolutional layer \cite{ECCV2016:UMDC}
with kernel size $1\times1$ and stride 1 before we obtain image features from the last two branches.
Both the output feature dimension of first GAT layer and input feature dimension of second
layer are 300 and the output feature dimension of second GAT layer is 512. The number of groups of labels $m$ is set as 4. The regularization parameter $\lambda$ is set to 0.001.
The input image is resized to 448x448 for both training and testing.
We train our model on one Tesla V100-16GB GPU and set the batch size to 8.
For optimization, we apply SGD as optimizer with momentum 0.9 and weight decay 10$^{-4}$.
The initial learning rate is set to 0.01 and the learning rate decay by 0.1 each 30 epochs in total 90 epochs \footnote{Source codes and pre-trained models of our method are publicly available at https://jiunhwang.github.io/}.

The evaluation metrics we used in our experiments include mean average precision (mAP) over all categories, precision (CP, OP), recall (CR, OR), and F1 score (CF1, OF1).

\begin{table*}[t]
\centering
\caption{Comparison of mAP and AP (in \%) of our method and state-of-the-art methods on Pascal VOC2007 dataset where numbers in bold indicate the best performance and numbers underlined indicate the second performance.}
\label{table:voc}
\scriptsize
\resizebox{\textwidth}{30mm}{
\begin{tabular}{c|cccccccc | cc}
\hline
voc2007 & CNN-RNN& ResNet-101&AR&ML-GCN&	A-GCN&	F-GCN	&SSGRL	&SSFRL(pre)&	\textbf{MSRN}&	\textbf{MRSN(pre)}\\ \hline
areo & 96.7 & 99.5 & 98.6 & 99.5 & 99.4 & 99.5 & 99.5 & 99.7 & \textbf{100.0} & 99.7 \\
bike & 83.1 & 97.7 & 97.1 & 98.5 & 98.5 & 98.5 & 97.1 & 98.4 & 98.8 & \textbf{98.9} \\
bird & 94.2 & 97.8 & 97.1 & 98.6 & 98.6 & 98.7 & 97.6 & 98.0 & \textbf{98.9} & 98.7 \\
boat & 92.8 & 96.4 & 95.5 & 98.1 & 98.0 & 98.2 & 97.8 & 97.6 & \textbf{99.1} & \textbf{99.1} \\
bottle & 61.2 & 75.7 & 75.6 & 80.8 & 80.8 & 80.9 & 82.6 & 85.7 & 81.6 & \textbf{86.6} \\
bus & 82.1 & 91.8 & 92.8 & 94.6 & 94.7 & 94.8 & 94.8 & 96.2 & 95.5 & \textbf{97.9} \\
car & 89.1 & 96.1 & 96.8 & 97.2 & 97.2 & 97.3 & 96.7 & 98.2 & 98.0 & \textbf{98.5} \\
cat & 94.2 & 97.6 & 97.3 & 98.2 & 98.2 & 98.3 & 98.1 & 98.8 & 98.2 & \textbf{98.9} \\
chair & 64.2 & 74.2 & 78.3 & 82.3 & 82.4 & 82.5 & 78.0 & 82.0 & 84.4 & \textbf{86.0} \\
cow & 83.6 & 80.9 & 92.2 & 95.7 & 95.5 & 95.7 & 97.0 & 98.1 & 96.6 & \textbf{98.7} \\
table & 70.0 & 85.0 & 87.6 & 86.4 & 86.4 & 86.6 & 85.6 & \textbf{89.7} & 87.5 & 89.1 \\
dog & 92.4 & 98.4 & 96.9 & 98.2 & 98.2 & 98.2 & 97.8 & 98.8 & 98.6 & \textbf{99.0} \\
horse & 91.7 & 96.5 & 96.5 & 98.4 & 98.4 & 98.4 & 98.3 & 98.7 & 98.6 & \textbf{99.1} \\
motor & 84.2 & 95.9 & 93.6 & 96.7 & 96.7 & 96.7 & 96.4 & 97.0 & 97.2 & \textbf{97.3} \\
person & 93.7 & 98.4 & 98.5 & 99.0 & 98.9 & 99.0 & 98.8 & 99.0 & 99.1 & \textbf{99.2} \\
plant & 59.8 & 70.1 & 81.6 & 84.7 & 84.8 & 84.8 & 84.9 & 86.9 & 87.0 & \textbf{90.2} \\
sleep & 93.2 & 88.3 & 93.1 & 96.7 & 96.6 & 96.7 & 96.5 & 98.1 & 97.6 & \textbf{99.2} \\
sofa & 75.3 & 80.2 & 83.2 & 84.3 & 84.4 & 84.4 & 79.8 & 85.8 & 86.5 & \textbf{89.7} \\
train & 99.7 & 98.9 & 98.5 & 98.9 & 98.9 & 99.0 & 98.4 & 99.0 & 99.4 & \textbf{99.8} \\
tv & 78.6 & 89.2 & 89.3 & 93.7 & 93.7 & 93.7 & 92.8 & 93.7 & 94.4 & \textbf{95.3} \\ \hline
mAP & 84.0 & 89.9 & 92.0 & 94.0 & 94.0 & 94.1 & 93.4 & 95.0 & 94.9 & \textbf{96.0} \\
\hline
\end{tabular}}
\end{table*}

\begin{table*}[t]
\caption{Comparison of our method with state-of-the-art methods on MS-COCO dataset where numbers in bold indicate the best performance and numbers underlined indicate the second performance.}
\label{table:coco}
\centering \scriptsize
\begin{tabular}{c|lllllll}
\hline
Methods & {mAP} &{CP} & {CR} & {CF1} & {OP} & {OR} & {OF1}  \\ \hline
CNN-RNN\cite{CVPR2016:ULMIC} & - & {-} & {-} & {-} & {-} & {-} & {-} \\
ResNet-101\cite{ResNet} & 80.3 & {77.8} & {72.8} & {75.2} & {81.5} & {75.1} & {78.2} \\
ML-GCN\cite{CVPR2019:ML-GCN} & 83.0 & 84.0 & {72.8} & 78.0 & 84.7 & {76.2} & 80.2 \\
A-GCN\cite{PRL:2020:AGCN} & {83.1} & {84.7} & {72.3} & {78.0} & 85.6 & 75.5 & 80.3 \\
F-GCN\cite{CIKM2020:FGCN} & {83.2} & {85.4} & 72.4 & \underline{78.3} & 86.0 & {75.7} & {80.5}\\
SSGRL\cite{ICCV2019:ssgrl} & \textbf{83.8} & \textbf{89.9} & {68.5} & {76.8} & \textbf{91.3} & {70.8} & {79.7}\\
CMA\cite{AAAI2020:CMA} & \underline{83.4} & {83.4} & \underline{72.9} & {77.8} & \underline{86.8} & \underline{76.3} & \underline{80.9}  \\
MS-CMA\cite{AAAI2020:CMA} & \textbf{83.8} & {82.9} & \textbf{74.4} & \textbf{78.4} & 84.4 & \textbf{77.9} & \textbf{81.0} \\

\textbf{MSRN} & \underline{83.4} & \underline{86.5} & 71.5 & \underline{78.3} & {86.1} & {75.5} & {80.4}  \\ \hline\hline

Methods & {CP-3} & {CR-3} & {CF1-3} & {OP-3} & {OR-3} & {OF1-3} &\\ \hline
CNN-RNN\cite{CVPR2016:ULMIC}  & 59.3 & 52.5 & 55.7& 59.8 & 61.4 & 60.7 \\
ResNet-101\cite{ResNet} & 84.1 & 59.4 & 69.7& 89.1 & 62.8 & 73.6 \\
ML-GCN\cite{CVPR2019:ML-GCN} & {89.2} & 64.1 & 74.6& {90.5} & 66.5 & 76.7 \\
A-GCN\cite{PRL:2020:AGCN} & 89.0 & 64.2 & 74.6& {90.5} & 66.3 & 76.6 \\
F-GCN\cite{CIKM2020:FGCN}  & \underline{89.3} & 64.3 & {74.7} & {90.5} & 66.6 & 76.7\\
SSGRL\cite{ICCV2019:ssgrl} & \textbf{91.9} & 62.5 & 72.7 & \textbf{93.8} & 64.1 & {76.2} \\
CMA\cite{AAAI2020:CMA}  & 86.7 & 64.9 & 74.3 & \underline{90.9} & 67.2 & \underline{77.2} \\
MS-CMA\cite{AAAI2020:CMA} & 88.2 & \underline{65.0} & \underline{74.9} & 90.2 & \underline{67.4} & 77.1 \\
\textbf{MSRN} & {84.5} & \textbf{72.9} & \textbf{78.3}& 84.3 & \textbf{76.8} & \textbf{80.4} \\ \hline\hline
\end{tabular}
\end{table*}

\subsection{Experimental results}
\noindent\textbf{VOC2007} \cite{voc2007}: We compare our method with ResNet-101 \cite{ResNet}, CNN-RNN \cite{CVPR2016:ULMIC}, AR \cite{AAAI2018:RARL}, ML-GCN \cite{CVPR2019:ML-GCN}, FGCN \cite{CIKM2020:FGCN}, SSGRL \cite{ICCV2019:ssgrl}.
Following \cite{ICCV2019:ssgrl}, we also pretrain our model on the MS-COCO dataset.
The results of all the methods are shown in Table \ref{table:voc}.
We can see that the result of MSRN(pre) is 2\%, 1.9\%, and 1\% better than ML-GCN \cite{CVPR2019:ML-GCN}, FGCN \cite{CIKM2020:FGCN}, SSGRL \cite{ICCV2019:ssgrl} on mAP respectively.
It should be noted that the input image size of SSGRL(pre) is 576$\times$576 which is larger than ours.
Our model also achieves the best AP score on 17 categories.
The results definitely demonstrate the effectiveness of modeling multi-layered semantic representations.

\noindent\textbf{MS-COCO} \cite{coco2014}: The comparison results on MS-COCO dataset are shown in Table \ref{table:coco}.
We compare our method with ResNet-101 \cite{ResNet}, ML-GCN \cite{CVPR2019:ML-GCN}, FGCN \cite{CIKM2020:FGCN} and CMA \cite{AAAI2020:CMA}.
Our method can achieve 83.4\% mAP score which is in the first rank.
Our model achieves comparable performance with the state-of-the-art methods.
Specifically, MSRN wins the second place in terms of mAP, CP, and CF1.
For the results on top-3 labels, MSRN obtains the best performance in terms of CR, CF1, OR and OF1.

\noindent\textbf{NUS-WIDE} \cite{Nuswide2009} contains 269,648 images and 81 concepts. The dataset is split by following \cite{AAAI2020:CMA}.
We compare MSRN with CNN-RNN\cite{CVPR2016:ULMIC},
AT\cite{ICLR2017:AT}, S-CLs\cite{CVPR2018:KDWD} and CMA\cite{AAAI2020:CMA}.
As shown in Table \ref{table:nuswide}, our model achieves 0.1\% better than MS-CMA on mAP.
In addition, MSRN achieves the best performance in terms of CF1, OF1 and CF1-3.
Both our model and MS-CMA\cite{AAAI2020:CMA} extract image features from lower layers of CNNs,
but our model outperforms it by 0.1\%, 0.2\%, 0.2\% and 0.4\% in terms of mAP, CF1, OF1 and CF1-3, respectively.

\begin{table}[h] \centering \small
\caption{Comparion with state-of-the-art methods on NUS-WIDE dataset where numbers in bold indicate the best performance and numbers underlined indicate the second performance.}
\label{table:nuswide}
\begin{tabular}{c|ccccc}
\hline
Methods & mAP & CF1 & OF1 & CF1-3 & OF1-3 \\ \hline
CNN-RNN\cite{CVPR2016:ULMIC} & 56.1 & - & - & 34.7 & 55.2\\
AT\cite{ICLR2017:AT} & 57.6 & 55.2 & 70.3 & 51.7 & 68.8\\
S-CLs\cite{CVPR2018:KDWD} & 60.1 & 58.7 & 73.7 & 53.8 & \textbf{71.1} \\
CMA\cite{AAAI2020:CMA} & {60.8} & {60.4} & {73.3} & {55.5} & \underline{70.0} \\
MS-CMA\cite{AAAI2020:CMA} & \underline{61.4} & \underline{60.5} & \underline{73.8} & \underline{55.7} & {69.5} \\
\textbf{MSRN} & \textbf{61.5} & \textbf{60.7} & \textbf{74.0} & \textbf{56.1} & {69.5}\\ \hline
\end{tabular}
\end{table}

\noindent\textbf{Apparel} \footnote{www.kaggle.com/kaiska/apparel-dataset} is a
clothing dataset for multi-label image classification.
We test our model on it and make comparisons with ResNet-101 \cite{ResNet}, SSGRL \cite{ICCV2019:ssgrl} and ML-GCN \cite{CVPR2019:ML-GCN}.
In our experiment, we randomly select 50\% images from the dataset for training, and other 50\% images for testing.
The result in Table $4$ shows that our model achieves 99.65 mAP score.
Our model is 0.18\% better than ResNet-101 and 0.06\% better than the current best model on mAP.
Our model also achieves the best score on all metrics that we employ.
\begin{table}[h]
\caption{Comparison with state-of-the-art methods on Apparel dataset where numbers in bold indicate the best performance and numbers underlined indicate the second performance.}
\label{table:apparel}
\centering \small
\begin{tabular}{c|ccccc}
\hline
Methods & mAP & CF1 & OF1 & CF1-3 & OF1-3 \\ \hline
ResNet-101\cite{ResNet} & 99.47 & 97.51 & 97.78 & 97.51 & 97.78\\
SSGRL\cite{ICCV2019:ssgrl} & \underline{99.57} & \underline{97.77} & \underline{98.01} & \underline{97.77} & \underline{98.01}\\
ML-GCN\cite{CVPR2019:ML-GCN} & {99.56}  & 97.68 & 97.88 & 97.68 & {97.87} \\
\textbf{MSRN} & \textbf{99.65} & \textbf{98.21} & \textbf{98.36} & \textbf{98.21} & \textbf{98.36}\\ \hline
\end{tabular}

\end{table}

\subsection{Ablation Studies}
In this section, we perform ablation studies to evaluate
the effectiveness of different components of our framework.

\noindent\textbf{Label and Group Embeddings}: To verify the effectiveness of label and group embeddings,
we conduct experiments with three simplified versions of our proposed method MSRN,
i.e., label-E (only using label embedding), group-E (only using group embedding), and no LGE module.
The results shown in Table \ref{table:LGE} clearly indicate the effectiveness of label and group embeddings.
\begin{table}[h]
\caption{Comparison among different versions of MSRN.}
\label{table:LGE}
\centering \small
\begin{tabular}{c|cccc}
\hline
Setting  & no LGE & label-E & group-E & {MSRN}\\ \hline
mAP & 91.75 &94.42 & {94.20} & \textbf{94.85}\\ \hline
\end{tabular}
\end{table}

\noindent\textbf{Number of Branches}: As ResNet-101 contains four blocks, we conduct experiments to
validate whether the multi-branch architecture is better than the single-branch architecture and whether the model performs better with all branches.
The experimental results are shown in Table \ref{tabel:branch}.
We can find that the multi-branch architecture can improve at least 0.12\% compared to the single-branch architecture, and achieves the best performance with the last 3 branches.
\begin{table}[h]
    \caption{Comparison among different number of branches.}
    \label{tabel:branch}
    \centering \small
    \begin{tabular}{c|cccc}
    \hline
        Number of branches & last 1 & last 2 & last 3 & all 4 \\ \hline
        mAP & 94.53 & \underline{94.66} & \textbf{94.85} & 94.65 \\ \hline
    \end{tabular}
\end{table}

\subsection{Parameter Sensitivity}
In this section, we study the sensitivity of MSRN to two hyper-parameters, i.e., the number of groups of labels $m$ and the regularization parameter $\lambda$. Due to the limitation of space, we only present the analyses on VOC2007 dataset.
For the number of groups $m$, we conduct experiments of six different cases corresponding to 2, 4, 6, 8, 10 and 20, respectively, with $\lambda$ fixed as $10^{-3}$.
The experimental results in Table \ref{table:groups} show that the performance in terms of mAP is not much sensitive to $m$. For $\lambda$, we study the values from $\{10^{-1}, 10^{-2}, ..., 10^{-6}\}$. The results with different values of $\lambda$ are shown in Table \ref{table:groups}, which shows the performance is not much sensitive to $\lambda$.
\begin{table}[h]
\caption{Comparison among different values of $m$ and $\lambda$.}
\label{table:groups}
\centering \small
\begin{tabular}{c |cccccc}
    \hline
    $m$ & 2 & 4 & 6 &8 & 10 & 20 \\ \hline
    mAP & 94.59 & \textbf{94.85} & 94.66 & 94.56 & 94.63 & 94.48 \\
    \hline\hline
    $\lambda$ & $10^{-1}$& $10^{-2}$& $10^{-3}$& $10^{-4}$& $10^{-5}$& $10^{-6}$\\
    \hline
    mAP & 94.43 & 94.74 & \textbf{94.85} & 94.69 & 94.66 & 94.55 \\\hline
    \end{tabular}
\end{table}

\section{Conclusion and Future Work}
\label{sec:conclusion}
This paper proposes a novel Multi-layered Semantic Representation Network (MSRN) for multi-label image classification. MSRN for the first time considers both local semantics and global semantics of labels through modeling label correlations, and learns semantic representations of images at multiple layers of a convolutional neural network through an attention mechanism.
Extensive experiments show that MSRN outperforms many state-of-art
methods on VOC2007, MS-COCO, NUS-WIDE and Apparel datasets.
In the future, we will improve our method to explicitly utilize labels which exist but are unobservable due to lack of labeling efforts.

\section*{Acknowledgement}
This work is supported by NSFC: 61806005 and 61906003, and The University Synergy Innovation Program of Anhui Province:GXXT-2020-012.

\bibliographystyle{abbrv}
\bibliography{msrn}

\begin{thebibliography}{10}

\bibitem{ECCV2016:UMDC}
Z.~Cai, , Q.~Fan, R.~S. Feris, and N.~Vasconcelos.
\newblock A unified multi-scale deep convolutional neural network for fast
  object detection.
\newblock In {\em ECCV}, pages 354--370, 2016.

\bibitem{AAAI2018:OFRNN}
S.~Chen, Y.~Chen, C.~Yeh, and Y.~F. Wang.
\newblock Order-free rnn with visual attention for multi-label classification.
\newblock In {\em AAAI}, 2018.

\bibitem{AAAI2018:RARL}
T.~Chen, Z.~Wang, G.~Li, and L.~Lin.
\newblock Recurrent attentional reinforcement learning for multi-label image
  recognition.
\newblock In {\em AAAI}, 2017.

\bibitem{ICCV2019:ssgrl}
T.~Chen, M.~Xu, X.~Hui, H.~Wu, and L.~Lin.
\newblock Learning semantic-specific graph representation for multi-label image
  recognition.
\newblock In {\em ICCV}, pages 522--531, 2019.

\bibitem{CVPR2019:ML-GCN}
Z.~Chen, X.~Wei, P.~Wang, and Y.~Guo.
\newblock Multi-label image recognition with graph convolutional networks.
\newblock In {\em CVPR}, pages 5172--5181, 2019.

\bibitem{Nuswide2009}
T.-S. Chua, J.~Tang, R.~Hong, H.~Li, Z.~Luo, and Y.~Zheng.
\newblock Nus-wide: a real-world web image database from national university of
  singapore.
\newblock In {\em ICIVR, ACM}, 2009.

\bibitem{voc2007}
M.~Everingham, L.~Gool, C.~K. Williams, J.~Winn, and A.~Zisserman.
\newblock The pascal visual object classes (voc) challenge.
\newblock {\em IJCV}, 88(2):303--338, 2010.

\bibitem{DenseNet}
H.~Gao, Z., M.~Liu, W.~Laurens, and Q.~Kilian.
\newblock Densely connected convolutional networks.
\newblock In {\em CVPR}, pages 4700--4708, 2017.

\bibitem{Chest-X-rays}
Z.~Y. Ge, D.~Mahapatra, S.~Sedai, R.~Garnavi, and R.~Chakravorty.
\newblock Chest x-rays classification: A multi-label and fine-grained problem.
\newblock {\em arXiv}, 2018.

\bibitem{ResNet}
K.~He, X.~Zhang, S.~Ren, and J.~Sun.
\newblock Deep residual learning for image recognition.
\newblock In {\em CVPR}, pages 770--778, 2016.

\bibitem{ICLR2018:MLAtt}
S.~Jetley, N.~A. Lord, N.~Lee, and P.~H.~S. Torr.
\newblock Learn to pay attention.
\newblock In {\em ICLR}, 2018.

\bibitem{CVPR2016:ODVT}
K.~Kang, W.~Ouyang, H.~Li, and X.~Wang.
\newblock Object detection from video tubelets with convolutional neural
  networks.
\newblock In {\em CVPR}, pages 817--825, 2016.

\bibitem{ICLR2017:HPLBP}
J.~Kim, K.~On, J.~Kim, J.~Ha, and B.~Zhang.
\newblock Hadamard product for low-rank bilinear pooling.
\newblock 10 2016.

\bibitem{CVPR2018:ZSL}
C.~Lee, W.~Fang, C.~Yeh, and Y.~F.Wang.
\newblock Multi-label zero-shot learning with structured knowledge graphs.
\newblock In {\em CVPR}, pages 1576--1585, 2018.

\bibitem{PRL:2020:AGCN}
Q.~Li, X.~Peng, Y.~Qiao, and Q.~Peng.
\newblock Learning label correlations for multi-label image recognition with
  graph networks.
\newblock {\em PRL}, 138:378--384, 2020.

\bibitem{coco2014}
T.~Lin, M.~Maire, S.~Belongie, L.~Bourdev, R.~Girshick, J.~Hays, P.~Perona,
  D.~Ramanan, C.~L. Zitnick, and P.~Dollr.
\newblock Microsoft coco: Common objects in context, 2014.

\bibitem{CVPR2018:KDWD}
Y.~Liu, L.~Sheng, J.~Shao, J.~Yan, S.~Xiang, and C.~Pan.
\newblock Multi-label image classification via knowledge distillation from
  weakly supervised detection.
\newblock In {\em ACM MM}, pages 700--708, 2018.

\bibitem{ICME2018:RCDH}
C.~Ma, Z.~Chen, J.~Lu, and J.~Zhou.
\newblock Rank-consistency multi-label deep hashing.
\newblock In {\em ICME}, pages 1--6, 2018.

\bibitem{Vgg}
K.~Simonyan and A.~Zisserman.
\newblock Very deep convolutional networks for large-scale image recognition.
\newblock In {\em ICLR}, 2015.

\bibitem{ICLR2018:GAT}
P.~Veli{\v{c}}kovi{\'{c}}, G.~Cucurull, A.~Casanova, A.~Romero, P.~Li{\`{o}},
  and Y.~Bengio.
\newblock {Graph Attention Networks}.
\newblock In {\em ICLR}, 2018.

\bibitem{CVPR2020:PRID}
D.~Wang and S.~Zhang.
\newblock Unsupervised person re-identification via multi-label classification.
\newblock In {\em CVPR}, pages 10978--10987, 2020.

\bibitem{CVPR2016:ULMIC}
J.~Wang, Y.~Yang, J.~Mao, Z.~Huang, C.~Huang, and W.~Xu.
\newblock {CNN-RNN:} {A} unified framework for multi-label image
  classification.
\newblock In {\em CVPR}, pages 2285--2294, 2016.

\bibitem{AAAI2020:KSSNet}
Y.~Wang, D.~He, F.~Li, X.~Long, Z.~Zhou, J.~Ma, and S.~Wen.
\newblock Multi-label classification with label graph superimposing.
\newblock In {\em AAAI}, pages 12265--12272, 2020.

\bibitem{CIKM2020:FGCN}
Y.~Wang, Y.~Xie, Y.~Liu, K.~Zhou, and X.~Li.
\newblock Fast graph convolution network based multi-label image recognition
  via cross-modal fusion.
\newblock In {\em CIKM}, pages 1575--1584, 2020.

\bibitem{ICCV2017:RDAR}
Z.~Wang, T.~Chen, G.~Li, R.~Xu, , and L.~Lin.
\newblock Multi-label image recognition by recurrently discovering attentional
  regions.
\newblock In {\em ICCV}, pages 464--472, 2017.

\bibitem{CVPR2016:EBBA}
H.~Yang, T.~Zhou, Y.~Zhang, B.~Gao, J.~Wu, and J.~Cai.
\newblock Exploit bounding box annotations for multi-label object recognition.
\newblock In {\em CVPR}, pages 280--288, 2016.

\bibitem{ICCV2015:MSR}
S.~Yang and D.~Ramanan.
\newblock Multi-scale recognition with dag-cnns.
\newblock In {\em ICCV}, pages 1215--1223, 2015.

\bibitem{NIPS18:Diffpool}
R.~Yin, J.~You, C.~Morris, X.~Ren, W.~L. Hamilton, and J.~Leskovec.
\newblock Hierarchical graph representation learning with differentiable
  pooling.
\newblock In {\em NIPS}, pages 4805--4815, 2018.

\bibitem{AAAI2020:CMA}
R.~You, Z.~Guo, L.~Cui, X.~Long, Y.~Bao, and S.~Wen.
\newblock Cross-modality attention with semantic graph embedding for
  multi-label classification.
\newblock In {\em AAAI}, volume~34, pages 12709--12716, 2020.

\bibitem{ICLR2017:AT}
S.~Zagoruyko and N.~Komodakis.
\newblock Paying more attention to attention: Improving the performance of
  convolutional neural networks via attention transfer.
\newblock In {\em ICLR}, 2017.

\bibitem{TKDE14:MLZhang:MLReview}
M.-L. Zhang and Z.-H. Zhou.
\newblock A review on multi-label learning algorithms.
\newblock {\em IEEE Trans. Knowl. Data Eng.}, 26(8):1819--1837, 2014.

\bibitem{CVPR2017:LSRIS}
F.~Zhu, H.~Li, W.~Ouyang, N.~Yu, and X.~Wang.
\newblock Learning spatial regularization with image-level supervisions for
  multi-label image classification.
\newblock In {\em CVPR}, pages 2027--2036, 2017.

\bibitem{ECCV2014:EB}
C.~L. Zitnick and P.~Doll{\'a}r.
\newblock Edge boxes: Locating object proposals from edges.
\newblock In {\em ECCV}, pages 391--405, 2014.

\end{thebibliography}

\end{document}